\documentclass[10pt, a4paper]{article}
\usepackage{lrec}
\usepackage{graphicx}
\usepackage{tabularx}
\usepackage{soul}
\usepackage{amsmath}

\usepackage{epstopdf}
\usepackage[latin1]{inputenc}

\usepackage{hyperref}
\usepackage{xstring}

\newcommand{\eg}{\textit{e.g., }}
\newcommand{\ie}{\textit{i.e., }}

\title{Building a Hebrew Semantic Role Labeling Lexical Resource\\
from Parallel Movie Subtitles}

\name{Ben Eyal, Michael Elhadad}

\address{Dept. of Computer Science, Ben-Gurion University of the Negev \\
         Beer Sheva, Israel \\
         bene@post.bgu.ac.il, elhadad@cs.bgu.ac.il\\}

\abstract{
We present a semantic role labeling resource for Hebrew built 
semi-automatically through annotation projection from English.
This corpus is derived from the multilingual OpenSubtitles dataset and includes
short informal sentences, for which reliable linguistic annotations
have been computed.  We provide a fully annotated version of the 
data including morphological analysis, dependency syntax and semantic role
labeling in both FrameNet and PropBank styles.  Sentences are aligned 
between English and Hebrew, both sides include full annotations and the explicit mapping from the English arguments to the Hebrew ones.
We train a neural SRL model on this Hebrew resource exploiting the pre-trained multilingual BERT transformer model, and provide the first
available baseline model for Hebrew SRL as a reference point. 
The code we provide is generic and can be adapted to other languages to bootstrap SRL resources.
\\ 
\newline 
 }

\begin{document}

\maketitleabstract

\section{Introduction}

Semantic role labeling (SRL) is the task that consists of annotating sentences with labels that answer questions such as ``Who did what to whom, when, and where?'', the answers to these questions are called ``roles.''

Two major SRL annotation schemes have been used in recent years, FrameNet \cite{baker1998berkeley}
and PropBank \cite{palmer2005proposition}. Research in SRL has generated fully-annotated corpora in various languages,
several CoNLL shared tasks, and many automatic SRL systems. 
Besides its important theoretical role, SRL has been found useful for information extraction \cite{christensen2010semantic,stanovsky2016supervised} and question answering \cite{fitzgerald2018largescale}.

In this work, we set out to contribute a new Hebrew corpus and lexical resources to further the advancement of the Hebrew FrameNet Project and to provide the first SRL resource covering both FrameNet and PropBank annotations \cite{hayoun2016hebrew}.  Our approach bootstraps Hebrew SRL
annotations by projecting predicted annotations from English to Hebrew aligned sentences.  To ensure that the resulting Hebrew annotations are reliable, we design confidence metrics on both the original English annotations and the projection method itself.

We describe previous work in multilingual FrameNet and cross-lingual SRL.
We then describe the large unannotated dataset we selected, OpenSubtitles, and the method we use to align documents and sentences between English (EN) and Hebrew (HE), and to project SRL annotations from EN to HE.  We present the dataset that is obtained as a result of this procedure, which includes aligned annotated sentences in EN and HE, with morphological analysis, dependency syntax and SRL in FrameNet and in PropBank formats.  Finally, we present baseline automatic SRL systems in HE trained on the generated data. 

Our main contribution in the paper is the description of an end to end method to bootstrap SRL datasets in low-resource settings through alignment, prediction and projection.  We provide full source code and datasets for the described method \url{https://github.com/bgunlp/hebrew_srl}.  The resulting dataset is the first available SRL resource for Hebrew.

\subsection{FrameNet}

This work is part of a larger effort to develop a Hebrew FrameNet \cite{hayoun2016hebrew}.  FrameNet is an annotation scheme and a lexical database inspired by frame semantics \cite{fillmore1976frame}.
Fillmore presented the concept of a `semantic frame' as
\begin{quote}
    A system of concepts related in such a way that to understand any one of them you have to understand the whole structure in which it fits.
\end{quote}
\newcite{baker1998berkeley} introduced a linguistic resource called FrameNet, on which work is ongoing to this day. 
The purpose of FrameNet is to realize the idea of frame semantics in English, by building a lexical database of annotated examples of how various words are used in actual texts, grouped by semantic frame.

The project defines a formal structure for semantic frames, and various relationships among them \newcite{ruppenhofer2016framenet}. The FrameNet lexical database is comprised of the following elements, which defines 
useful terminology for all SRL resources:

\begin{itemize}
\item[] \textbf{Frames.} Each frame contains a list of frame evoking words, such as ``bought'' in the sentence ``John bought a car from Jane.''
        These are known as Frame Evoking Elements (FEEs) or Lexical Units (LUs).
        Additionally, each frame defines a list of participants and a list of constraints
        on and relationships between these participants.
        The participants are called Frame Elements (FEs).
\item[] \textbf{Lexical Units.} Formally, an LU is a word lemma paired with a coarse part-of-speech tag
        and is unique within its frame.
        For example, both the words \textbf{bought} and \textbf{buying} are 
        represented by the LU \textbf{buy.v} in the \textsc{Commerce\_buy} frame.
        
        In the FrameNet formalism, LUs can have almost any part-of-speech tag.
        As an example consider the noun \textbf{purchase} (as in ``a purchase was made''),
        which is one of the LUs in the \textsc{Commerce\_buy} frame.
        Verbs, however, are the most common LUs.  In this work we only
        consider verbal LUs.
        
        A \textbf{target} is any constituent which evokes a frame.
        It is the instance of an LU in a given piece of text.
        For example, both \textbf{buying} and \textbf{bought} could be targets
        which evoke the \textsc{Commerce\_buy} frame and they represent the LU \textbf{buy.v}.
\item[] \textbf{Frame Elements.} FrameNet classifies frame elements in terms of how central they are to the frame.
        The two primary levels of centrality are labeled ``core'' and ``peripheral''.
        
        An FE is classified as a \emph{core} FE if it instantiates a conceptually
        necessary component of a frame, while making a frame unique.
        For example, in the \textsc{Commerce\_buy} frame,
        the FEs \textbf{Buyer} and \textbf{Goods} are considered core elements,
        while the FE \textbf{Money} (representing the thing given in exchange for the goods) is not.
        
        In determining which FEs are core in their frame, a few formal properties are considered which may provide evidence for core status:
        
        \begin{itemize}
        \item When an element must be explicitly expressed, it is core. For instance, \textit{resemble} requires two entities to compare.
        \item If an element receives a definite interpretation when omitted, it is core. For example, the sentence ``John arrived.''
        is incomplete if the goal location at which John arrived cannot be inferred from context.
        \end{itemize}
        
        FEs that do not introduce additional, independent or distinct events from the main reported event are classified as \emph{peripheral}.
        Peripheral FEs mark notions such as time, place, manner, degree of the main event represented by the frame.
\end{itemize}

FrameNet in English is currently at version 1.7 which consists of 1,221 frames, 13,572 LUs, and 11,428 FEs.  For training purposes, there are 10,147 sentences in 107 fully-annotated texts.


\subsection{FrameNet in other Languages}

A great effort is made to expand FrameNet to other languages.
As of today, FrameNets have been developed in
Finnish \cite{linden2017finnfn},
Spanish \cite{subirats2003surprise},
German \cite{burchardt2006salsa},
Japanese \cite{ohara2004japanese},
Chinese \cite{you2005building},
Korean \cite{nam2014korean},
Brazilian Portuguese \cite{torrent20133},
Swedish \cite{borin2010past},
French \cite{candito2014developing},
Danish \cite{bick2011framenet},
Polish \cite{zawislawska2008framenet},
Italian \cite{tonelli2008frame},
Slovenian \cite{lonneker2008new},
and of course, Hebrew \cite{petruck2005towards,hayoun2016hebrew}.

The methods to develop FrameNets presented in these papers are quite similar: 
assume the universality of the English frame inventory, and under that assumption,
tag, either manually or semi-automatically, sentences in the desired language using this inventory.
Almost every language has its specific corner cases where the English frames
are insufficient. In those cases, usually new frames are created specifically for the language in question. 
An example of a corner case of Hebrew is
multi-word lexical units like \textit{give up} and \textit{turn in} -
in Hebrew, these LUs might not appear as contiguous words. This case
was solved by allowing annotation of discontinuous units.

\subsection{Annotation Projection}

The idea of annotation projection is used in most research addressing
cross-lingual SRL and linguistic annotation
\cite{yarowsky-etal-2001-inducing,pado2007cross,pado2009cross,van-der-plas-etal-2011-scaling}. In this line of work, we start with a resource-rich source language, usually English, and a parallel corpus of English and a resource-poor target language. The English sentences are annotated using  an automatic tool, a semantic role labeling model in our case, and using word alignment, the annotations are projected to the target language. 
This process is called ``direct projection.''   
The idea of ``filtered projection'' is introduced in \newcite{akbik-etal-2015-generating}: only alignments which satisfy the suggested filters are kept, while the rest are discarded. Such filters include verb filter (discard sentences where the predicate is not aligned to a verb), and translation filter (discard sentences where the aligned predicate is not a translation of the source predicate).  We find in this work that filtered projection is essential to produce reliable annotations in Hebrew.

In the rest of the paper, we present related work in the field of cross-lingual SRL annotation projection and the method we used to construct a Hebrew SRL dataset starting from a large parallel corpus of English/Hebrew sentence pairs.  We eventually train a Hebrew SRL system on the data we produce and report on its performance on the automatically generated data.

\section{Related Work}

A long tradition of research has investigated how to create FrameNets and PropBanks in multiple languages using annotation projection. 

For FrameNet, \newcite{pado2009cross} present an approach to automatically create a German FrameNet which formulates the search for a semantic alignment (an alignment in which each pair of aligned constituents are semantically equivalent) as an optimization problem on a bipartite graph.  They report an F1 measure of 56.0 for Frame Elements prediction in German.
\newcite{annesi2010cross} set out to create an Italian FrameNet by applying a Hidden Markov Model to project annotations from English to Italian, with 
an F1 measure of 60.3 for Frame Elements prediction in Italian.

\newcite{van-der-plas-etal-2011-scaling} use the PropBank-style annotations for their system (PropBank resources enjoy more annotated data than FrameNet). Their target language is French, and the pipeline consists of training a French syntactic parser, transfer the English roles to French via word alignments, and finally, train a French joint syntactic-semantic parser on the French syntactic and semantic annotations. They report argument labeling performance in French of F1=65.

\newcite{akbik-etal-2015-generating} contributed PropBanks in seven languages (Arabic, Chinese, French, German, Hindi, Russian, Spanish) using filtered annotations projection\footnote{\url{https://github.com/System-T/UniversalPropositions}}.  The reported argument labeling performance ranges between 65.0 (Arabic) and 82.0 (Chinese).  Filtering projections significantly improves on previous work: for example, for French, the reported performance for argument exact match is F1=80.0 compared to the 65.0 reported above.

More recently, \newcite{aminian2019crosslingual} presented a deep bidirectional character-level LSTM encoder-decoder model which uses annotation projection to label new languages, without using any syntactic features such as part-of-speech tags and dependency trees.  This end to end model improves slightly over the baseline results presented in \newcite{akbik-etal-2015-generating}.

A recent entry to the list of automatic cross-lingual SRL systems is by \newcite{daza-frank-2019-translate}, which does not use annotation projection in order to label semantic roles in a different language, but rather learns to simultaneously translate and label the input sentence.
The reported performance F1=77.2 for German and F1=72.4 for French.

\section{Data Collection}

We adopt in this work the method of filtered projection of \newcite{akbik-etal-2015-generating}, and present a pipeline 
of filters that we introduce to control the quality of the 
generated dataset.  We start from a noisy collection of aligned 
documents in large quantities (23.7M aligned sentences) in the OpenSubtitles 2016 dataset \cite{LISON16.947}.  

We apply filters of different types on this data: starting with language identification, we push the data in Hebrew and English through automatic NLP annotation (POS, dependency parsing and SRL for English), we compute word-level alignment. We then design a method to rank annotated sentence pairs in terms of syntactic plausibility (that is how likely it is that the English and Hebrew parse trees will project onto each other in a clean manner).  
Modern Hebrew is characterized by a rich morphological system which impacts annotation schemas, especially because function words are often agglutinated together with the content words they modify.  
In terms of syntax, Modern Hebrew is mainly an SVO word order with no explicit case marking (except for pronouns).  Personal pronouns are often skipped when the verb morphological inflection provides clear cues (\textit{e.g.,} ``ahav-ti / I liked'' instead of 
``ani ahav-ti'').

\begin{figure*}
\centering
\includegraphics[width=\textwidth]{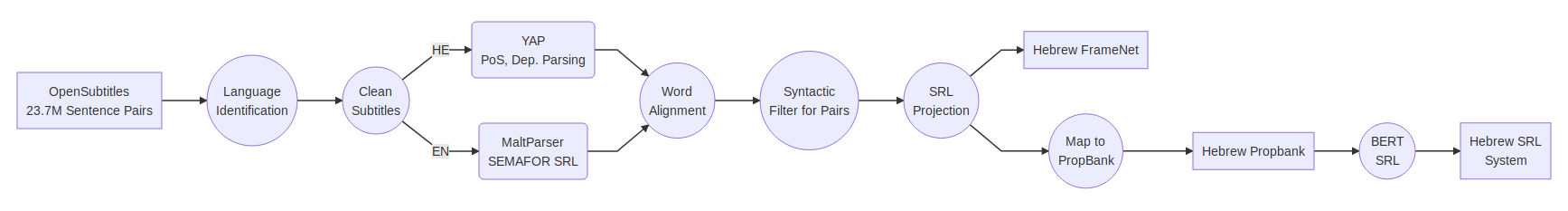}
\caption{Overall SRL dataset production pipeline}
\label{fig:process}
\end{figure*}

The overall data annotation process is summarized in Fig.~\ref{fig:process}.  We describe the steps of this process in the following paragraphs.

\subsection{Filtering Noisy Subtitles}

The OpenSubtitles 2016 dataset \cite{TIEDEMANN12.463} provides aligned data in 62 languages from movie subtitles.  It specifically includes 23.7 million English-Hebrew sentence pairs. Due to the nature of the dataset, some of the pairs are very noisy. Examples of such noise include special Unicode characters unrelated to the text, \eg musical note symbol to let hearing impaired viewers know that a song is playing, joined words (either typos or OCR artefacts) such ``Amanonce'' instead of ``A man once,'' and pairs in which the Hebrew part is actually found in English, not translated.

Given the large number of available sentence pairs, we filter noisy pairs in the dataset. The removal of noise consists of running fastText's language detection model \cite{joulin2016bag,joulin2016fasttext} on each pair, and keeping only English-Hebrew pairs.  After this first filtering, we have 22.4M sentence pairs. 

\subsection{Hebrew Preprocessing}

The importance of Hebrew preprocessing in our work is twofold:
(a) we use features from the dependency parse tree later on in the pipeline,
and (b) it acts as a quality gate for our data.

The preprocessing pipeline consists of morphological analysis, morphological
disambiguation, and dependency parsing, all using the YAP system \cite{more-etal-2019-joint}. In Hebrew, common words including prepositions, conjunctions and articles are written in an aggregated manner. For example, the written form for the phrase ``in the house'' (be ha bayit) will be written as a single token (babayit).  For the goal of SRL, it is particularly relevant to segment such compound tokens so that prepositions and conjunctions can be properly aligned with their English counterparts and projected with the corresponding English sentences.

Before segmentation, the data includes 118M tokens from 2.4M word types.  We predict that after segmentation, the number of tokens will rise, but the number of word types will be significantly lower. Indeed, after segmentation
we are left with 188M tokens and 0.89M word types.
This is a surprising number, as one would expect the number of
types in Hebrew to be approximately 200-300K, not 890K. A quick check
shows that 782K word types appear less than ten times in the entire dataset,
leaving us with a more reasonable vocabulary of 112K word types.

Naturally, segmentation also affects sentence length, with 5.13 words before
segmentation, and 8.17 after. Both of these numbers indicate
that the sentences in the dataset are very short. The genre that this dataset represents is informal spoken language, with a high prevalence of personal pronouns (I, you, he, she) and modals (can, should). The three most prominent parts-of-speech are Nouns with 22M instances, personal pronouns with 13M instances, and verbs with 13M instances (including modals).

Dependency parse tree have an average depth of 2.96, which confirms that the average sentence is very short and simple, containing most often a single predicate with its arguments.  We discuss below that as part of the overall preprocessing pipeline, we filter sentences that are either too short or too shallow to be of interest for SRL purposes.

\subsection{English Preprocessing}

We apply the pre-trained SEMAFOR system \cite{das2014frame} on English sentences to obtain syntax and SRL annotations.  
As part of SEMAFOR's  pipeline, pre-processing is done on the input sentences using MaltParser \cite{nivre2007maltparser} pre-trained on
sections 02-21 of the WSJ section of the Penn Treebank.

On the English side, the dataset contains 148M tokens consisting
of 1.5M types, with sentence length averaging at 6.45.
41M of them are nouns, 36M personal pronouns,
and 23M verbs. Compare the 36M personal pronouns in English with 13M in Hebrew.  This is a known syntactic aspect in Hebrew where personal pronouns are often unmarked and understood from the morphological inflection of the main verb.  Such a mismatch between EN and HE challenges the methods for constituent alignment.  The depth of the dependency parse tree in English is slightly smaller than in Hebrew, averaging at 2.44.

\subsection{Semantic Role Labeling}

The actual output from SEMAFOR consists of 55M instantiated frames across the entire dataset, but only 784 unique frames were evoked out of the 1,020 frames available in FrameNet v1.5, on which SEMAFOR is trained. 
On average, each sentence contains 2.4 frames and 1.17 frame elements per frame.  
Before filtering, the most general frame elements are also the most frequent - 3.9M instances of \textsc{Entity}, 3.3M instances of \textsc{Agent}, and 3.1M instances of \textsc{Theme}.

After the complete filtering pipeline, however, the most frequent frames are \textsc{Statement}, \textsc{Arriving}, \textsc{Becoming}, \textsc{Motion} and \textsc{Killing} (reflecting the violent nature of movies) and the most frequent frame elements are \textsc{Theme}, \textsc{Speaker}, \textsc{Message}, \textsc{Agent} and \textsc{Goal}.

\subsection{Computing Word Alignments}

The original data obtained from OpenSubtitles provides aligned sentence pairs.
In order to project SRL annotations from English to Hebrew, we must compute 
word alignment.  To this end, we use the fast\_align method of \newcite{dyer-etal-2013-simple} on the set of aligned sentences.  The output provides a directional mapping from English tokens in a sentence to the corresponding tokens in the associated Hebrew sentence.  This alignment is not necessarily one to one.

The word alignment process identified 181M word pairs. 
On average, a single English token is mapped to 1.25 Hebrew after segmentation.
Of these 181M pairs, 115M are one-to-one alignments, 29M are cases of 1-n alignments.
In the case of a 1-n alignment, the single English token is mapped to multiple Hebrew tokens and these 29M cases cover 66M pairs.   

Within the 181M collected pairs, we identify 17.6M distinct word pairs (for the full vocabulary of about 1.5M distinct words in English). Given the fact that Hebrew has rich morphology, the average rate of distinct Hebrew tokens associated to the same English token is as expected (each English form can be mapped to multiple inflected Hebrew forms in addition to the natural expected lexical diversity).

Manual inspection indicated that when one English word is mapped to many Hebrew tokens (1-3 or more), it is an indication of poor data quality. This feature was, therefore, added in the pair quality filter described below.

\subsection{Span Projection}

On the basis of the word alignment, it is possible to project the English SRL annotations onto the Hebrew sentence.  The projection process traverses the dependency syntax tree recursively: it identifies the head word of an annotated span in English, maps it to the word aligned in Hebrew, find its subtree in the Hebrew side, and annotates it accordingly.

This simple projection method cannot work well in the presence of noisy word alignments.  For example, if the head word of a span in English is aligned with zero or more than one word in Hebrew, the procedure will not produce aligned constituents.  Given the abundance of data, we simply skip these problematic alignment cases.  
Out of the 22M sentence pairs, 6.8M annotated pairs are left after this projection filter.

\subsection{Filtering Sentence Pairs}

After preprocessing both English and Hebrew, we filter the 6.8M sentence pair in a way that reflects our confidence in the alignment between the annotated English and Hebrew syntactic structures.

\begin{figure*}
\centering
\includegraphics[width=\textwidth]{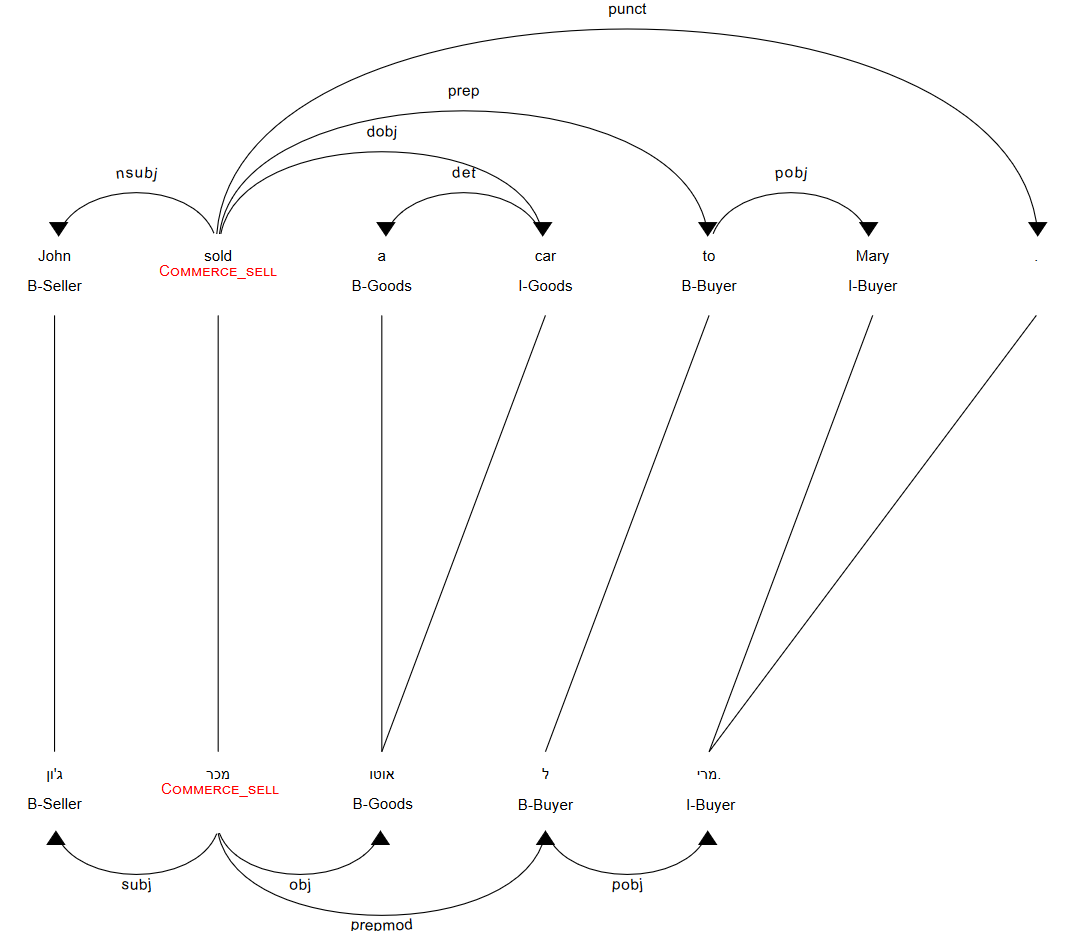}
\caption{Visualization of the English to Hebrew SRL Projection}
\label{fig:tree}
\end{figure*}

To this end, we built a graphical user interface to visualize the word alignment,
dependency tree of both English and Hebrew, the English SRL annotations, and the Hebrew SRL projections as shown in Fig.~\ref{fig:tree}.  This diagram summarizes all the NLP annotations available on both languages and the mapping between the SRL spans in a synthetic manner. 

We manually annotated 124 sentences as one of the following six options:
\begin{itemize}
\item Error in sentence alignment *
\item Poor translation *
\item Error in word alignment
\item Poor syntactic parsing
\item Poor frame parsing
\item Good
\end{itemize}
The items marked with an asterisk are problems in the dataset itself, \ie better tools cannot improve these sentences' quality.

Using these annotations and random resampling, we train a linear classifier to automatically annotate other sentences.
We annotate a new sentence if the classifier gives it more than 80\% of being ``Good''.
For training, we use the following features:
\begin{itemize}
\item Sentence lengths (English/Hebrew)
\item English-Hebrew sentence length ratio
\item Number of frames in the English sentence
\item Number of one-to-one word alignments, \ie which align one English word to one Hebrew word
\item Number of one-to-many word alignments
\item Dependency parse tree depth (English/Hebrew)
\end{itemize}

We interpret the prediction returned by the classifier as a score indicating the quality of the SRL match between the English and Hebrew sentences.  Higher scores are associated with longer sentences (for example, with a score over 0.80, the average sentence length is 19.74 words vs. 8.2 for the whole data) with good sentence and word alignments when verified manually.  

\begin{figure*}
\centering
\includegraphics[width=\textwidth]{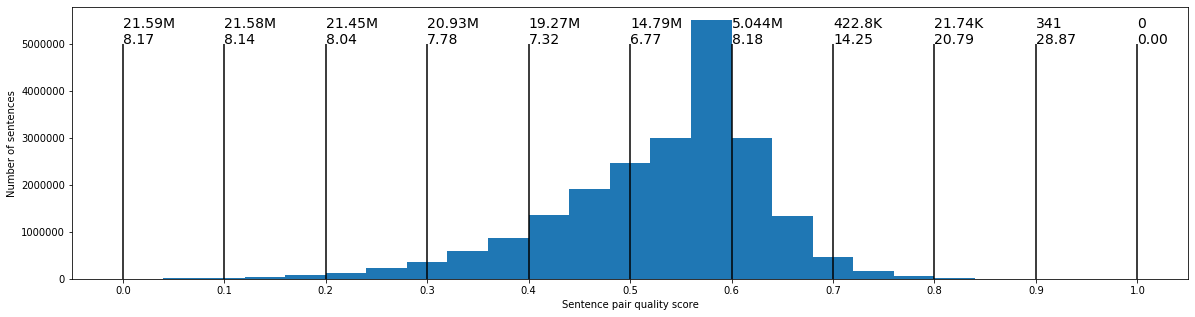}
\caption{Histogram of sentence pairs scores and sentence length}
\label{fig:histo}
\end{figure*}

By adjusting the threshold on the reliability score produced by this classifier, we can control how many sentence pairs we use in the training phrase of the pipeline.
The distribution of sentence pairs scores and their length is shown in Fig.~\ref{fig:histo}, where the numbers on top indicate for each bin, the number of sentence pairs with a score above a given threshold and the average length of the Hebrew sentences with a score above this threshold.  For example, with a threshold above 0.70, there are 422.8K sentence pairs with an average length of 14.25 segmented Hebrew tokens.  

\subsection{Mapping to PropBank}

Given a Hebrew SRL resource, we can train a Hebrew SRL system.
Because there are no available FrameNet parsers for languages other than English, we decided to map the projected FrameNet annotations we collected to PropBank-style annotations in order to make use of existing automatic SRL systems. In this mapping, we only consider annotations with FEs marked as ``Core'' by FrameNet.  We map these Core FEs to PropBank arguments by relying on the order in which they appear in the FrameNet frame definition, and map them accordingly to \texttt{ARG0}, \texttt{ARG1}, etc. For example, the \textsc{Commerce\_buy} frame has two core FEs: Buyer and Goods. In the sentence ``Abby bought a car from Robin for \$5,000'', ``Abby'' is the Buyer and ``a car'' is the Goods. This sentence will be labeled as ``\( [\text{Abby}_{\text{\texttt{ARG0}}}] \text{\textbf{ bought }} [\text{a car}_{\text{\texttt{ARG1}}}] \text{ from Robin for \$5,000} \)''. 

To apply this mapping to PropBank, we only keep sentences that include only Core Frame Elements\footnote{Mapping non-core elements such as space or time between FrameNet and PropBank is more interesting semantically, but we found it challenging to perform this mapping automatically.  Noisy mapping would increase the risk of producing bad PropBank annotations, and we end up with a good enough number of sentences with the more stringent filter.}.
Upon further filtering with this criterion, we are left with 423K annotated sentences.

We sample from these sentences a training set of 240K sentences, 30K for development and 30K for test.  
The statistics of the Hebrew PropBank dataset are shown in Table~\ref{ds-stats}.  

\begin{table*}
\centering
\caption{Dataset Statistics. (S) stands for segmented Hebrew words, (U) for unsegmented, ASL is ``average sentence length''}
\label{ds-stats}
\begin{tabular}{|l|l|l|l|l|l|l|l|} 
\hline
Fold        & \#sentences & \#tokens (S) & \#types (S) & ASL (S) & \#tokens (U) & \#types (U) & ASL (U)  \\ 
\hline
Train       & 240K     & 2.2M    & 60,288      & 9.2     & 1.8M    & 95,835      & 7.6      \\ 
\hline
Development & 30K      & 277K      & 20,633      & 9.2     & 228K      & 28,714      & 7.6      \\ 
\hline
Test        & 30K      & 277K      & 20,664      & 9.2     & 229K      & 28,770      & 7.6      \\
\hline
\end{tabular}
\end{table*}

\section{Experimental Results}

\subsection{Model}

We train a Hebrew PropBank semantic role labeler on the dataset we constructed to provide a baseline. 

To this end, we use AllenNLP's \cite{Gardner2017AllenNLP} implementation of the model by \newcite{shi2019simple}, which only uses BERT \cite{devlin2018bert} for argument identification and classification. The specific BERT model we adopt is the multilingual cased version of \texttt{BERT-Base} which was trained, among other languages, on Hebrew. Hyper-parameters are as specified in AllenNLP's configuration file\footnote{
\url{https://github.com/allenai/allennlp/blob/master/training_config/bert_base_srl.jsonnet}}.  This model simply encodes the whole sentence using the BERT model and predicts the BIO tags for the arguments given the predicate.
A sentence is passed to BERT in the form \textit{[CLS] sentence [SEP] pred [SEP]} - for example, \textit{[CLS] Barack Obama went to Paris [SEP] went [SEP]}.  \newcite{shi2019simple} report a performance of F1=82.7 on argument identification on CoNLL 2009 in English.

\subsection{Evaluation}

As BERT works with WordPiece tokenization \cite{sennrich-etal-2016-neural}, we hypothesized that the segmented Hebrew words, \ie the phrase ``and from you / ve min ata'' is one Hebrew word which is segmented into three words, would perform worse, as BERT has not seen segmented sentences. To this end, we unsegment the sentences back to their original form (joining ``and from you / ve min ata'' back into one token ``vemimkha'') to see if this would improve performance. Table \ref{train-perf} shows that this hypothesis is incorrect, as training with both segmented and unsegmented tokens provides roughly equivalent performance.

\begin{table}
\centering
\caption{Training Performance}
\label{train-perf}
\begin{tabular}{l|l|l|l}
            & Precision & Recall & F1    \\ 
\hline
Segmented   & 0.65      & 0.63   & 0.64  \\ 
\hline
Unsegmented & 0.64      & 0.62   & 0.63 
\end{tabular}
\end{table}



\section{Conclusion}

We present the first available SRL resource in Hebrew, which is constructed through annotation projection from aligned English sentences.  The dataset is derived from the OpenSubtitles collection of movie subtitles.  The genre is of informal spoken language.  We designed a full pipeline to map a noisy collection of aligned sentences in English and Hebrew into an annotated SRL dataset in both FrameNet and PropBank styles in CoNLL 2009 format.  

In order to control the quality of the generated data, we introduce a set of filters, following the approach of filtered projection of \newcite{akbik-etal-2015-generating}.  We specifically verified the impact of the Hebrew rich morphology and complex word formation rules on the process.

We finally trained a neural SRL system in Hebrew as a baseline model, building on the BERT multi-lingual model. 

In future work, we intend to manually curate the generated data while enriching the Hebrew FrameNet database with annotated exemplar sentences.


Code, data and pre-trained models for Hebrew SRL are available at \url{https://github.com/bgunlp/hebrew_srl}.

\section{Acknowledgements}

This research was supported by the Lynn and William Frankel Center for Computer Science at Ben Gurion University.

\section{Bibliographical References}

\bibliographystyle{lrec}
\bibliography{paper}

\section*{Appendix}

Below are the inputs to the SRL projection for the English sentence ``John sold a car to Mary.''

\begin{figure}[h]
    \centering
    \includegraphics[scale=0.25]{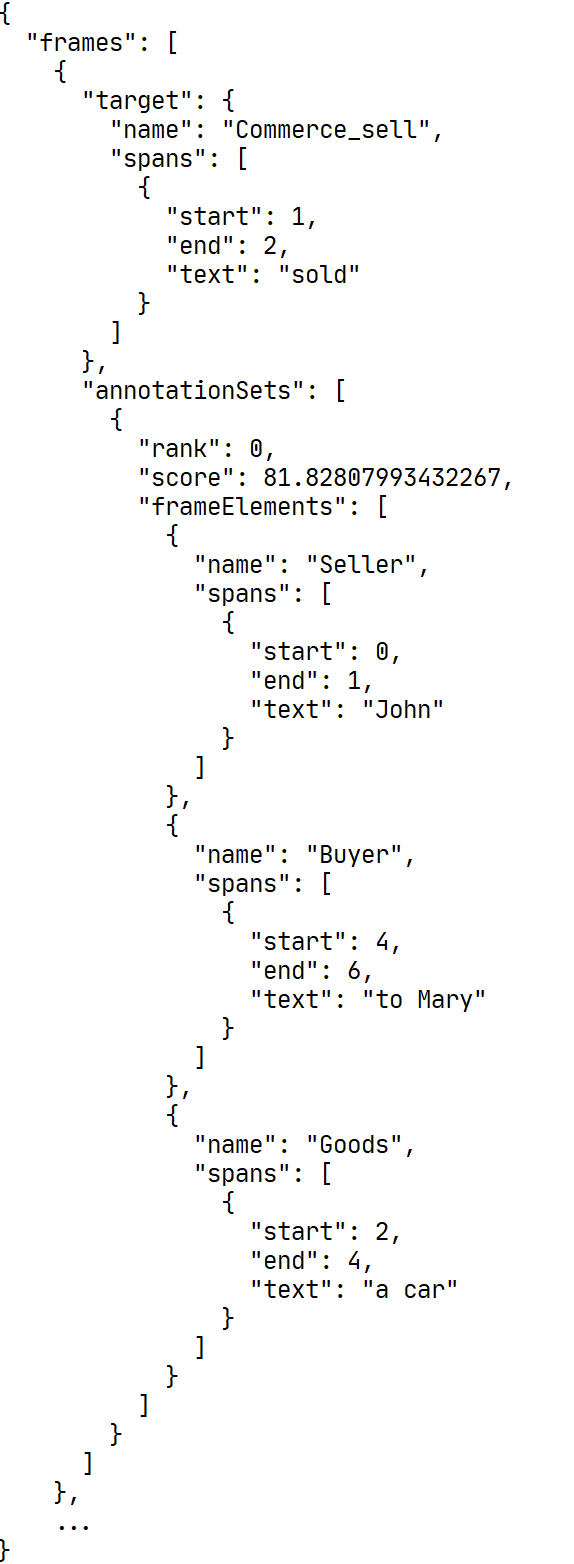}
    \caption{English SRL input}
\end{figure}

\begin{figure*}
    \centering
    \includegraphics[width=\textwidth]{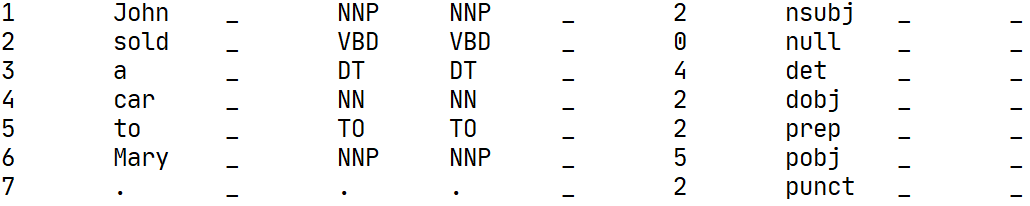}
    \caption{English input}
\end{figure*}

\begin{figure*}
    \centering
    \includegraphics[width=\textwidth]{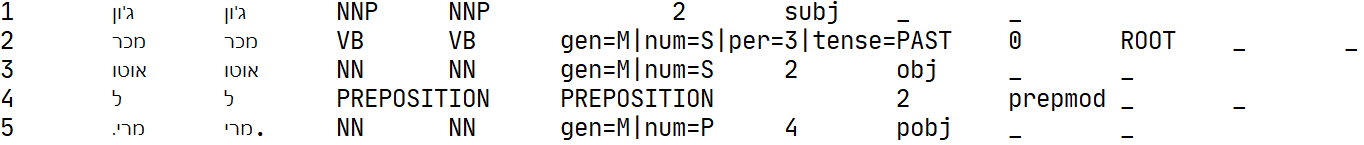}
    \caption{Hebrew input}
\end{figure*}

\begin{figure*}
    \centering
    \includegraphics[scale=0.5]{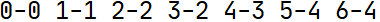}
    \caption{\texttt{fast\_align} word alignments}
\end{figure*}

\end{document}